\documentclass[letterpaper, 10 pt, conference]{ieeeconf}  

\IEEEoverridecommandlockouts                              

\usepackage{graphicx}
\usepackage{subcaption}
\usepackage{amsmath}
\usepackage{amssymb}
\usepackage{breqn}
\usepackage{verbatim}
\usepackage{balance}
\usepackage{color}

\graphicspath{{imgs/}}

\title{\LARGE \bf
Tool Substitution with Shape and Material Reasoning \\ Using Dual Neural Networks
}

\author{Nithin Shrivatsav$^{1}$, Lakshmi Nair$^{1}$ and Sonia Chernova$^{1}$
\thanks{$^1$ Georgia Institute of Technology, Atlanta, GA. Email: {\tt\small \{nithinshri,lnair3,chernova\}@gatech.edu}}}%

\begin{document}

\maketitle
\thispagestyle{empty}
\pagestyle{empty}

\begin{abstract}

This paper explores the problem of tool substitution, namely, identifying substitute tools for performing a task from a given set of candidate tools. We introduce a novel approach to tool substitution, that unlike prior work in the area, combines both shape \textit{and} material reasoning to effectively identify substitute tools. Our approach combines the use of visual and spectral reasoning using dual neural networks. It takes as input, the desired action to be performed, and outputs a ranking of the available candidate tools based on their suitability for performing the action. Our results on a test set of 30 real-world objects show that our approach is able to effectively match shape and material similarities, with improved tool substitution performance when combining both.


\end{abstract}

\section{INTRODUCTION}

Tool substitution addresses scenarios where a robot is tasked with accomplishing an action requiring a canonical tool, and in its absence, must find an alternative among candidates within its environment, e.g., the robot may have to perform the task of scooping beans, and in the case of a missing scoop/ladle, might use a mug instead. Such scenarios often require the robot to adapt knowledge of the canonical tool to the candidate tools now available to it and improves the resourcefulness of robots in unprecedented scenarios requiring creative use of available objects. Prior work has looked at tool substitution, reasoning primarily about visual properties such as shapes and sizes of the different parts of the candidate tools \cite{abelha2016model, schoeler2016bootstrapping}. However, they do not reason about material properties, which play an important role in identifying whether an object makes a good substitute for a missing tool, e.g., between a metal cup and a plastic cup, the metal cup would be a better substitute for a hammer. 

\begin{figure}[t]
	\centering
	\includegraphics[width=0.45\textwidth]{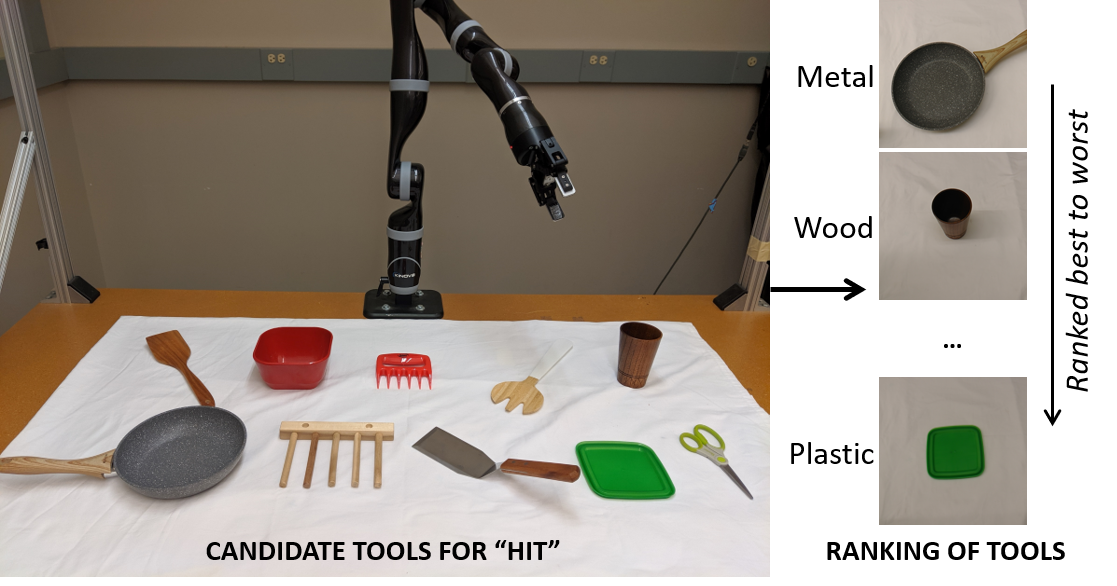}
	\captionsetup{width=\linewidth}
	\caption{Given an action to perform (e.g., `Hit') and candidate tools, our approach uses combined shape and material reasoning to output a ranking of the tools.} 
	\label{fig:Overview}
\end{figure}

In this work, we tackle the problem of tool substitution, and in contrast to prior work in the field, introduce a novel approach to reason about \textit{both shape and materials}, to perform efficient tool substitution. More specifically, given an action to perform, e.g., `hit', and a set of candidate tools available to the robot, our approach outputs a ranking of the candidate tools based on their appropriateness for performing the specified action (See figure \ref{fig:Overview}). Intuitively, our approach reasons about the degree to which a given point cloud shape and material are similar to that of canonical tools used for a given action, allowing us to differentiate and rank the candidates based on the similarity score. In this work, we use Dual Neural Networks\footnote{Also known as Siamese neural networks. We avoid using the term ``Siamese'', instead referring to such networks as Dual Neural Networks in our paper}, owing to their prior success in scoring similarity of two given inputs \cite{koch2015siamese, bromley1994signature}. Our work contributes the following:

\begin{itemize}
    \item A novel approach for matching similarity of \textit{shapes} of point clouds using dual neural networks;
    \item A novel approach for matching similarity of \textit{materials} based on spectrometer data using dual neural networks;
    \item A tool substitution framework that incorporates \textit{both} shape and material reasoning to rank the candidate tools. 
    
\end{itemize}

We demonstrate the efficiency of our tool substitution approach on a test set of 30 real-world objects, for performing six different actions.




\section{RELATED WORK}
In this section, we summarize some closely related work. 

\subsection{Tool Substitution}
\begin{figure}[t]
	\centering
	\includegraphics[width=0.43\textwidth]{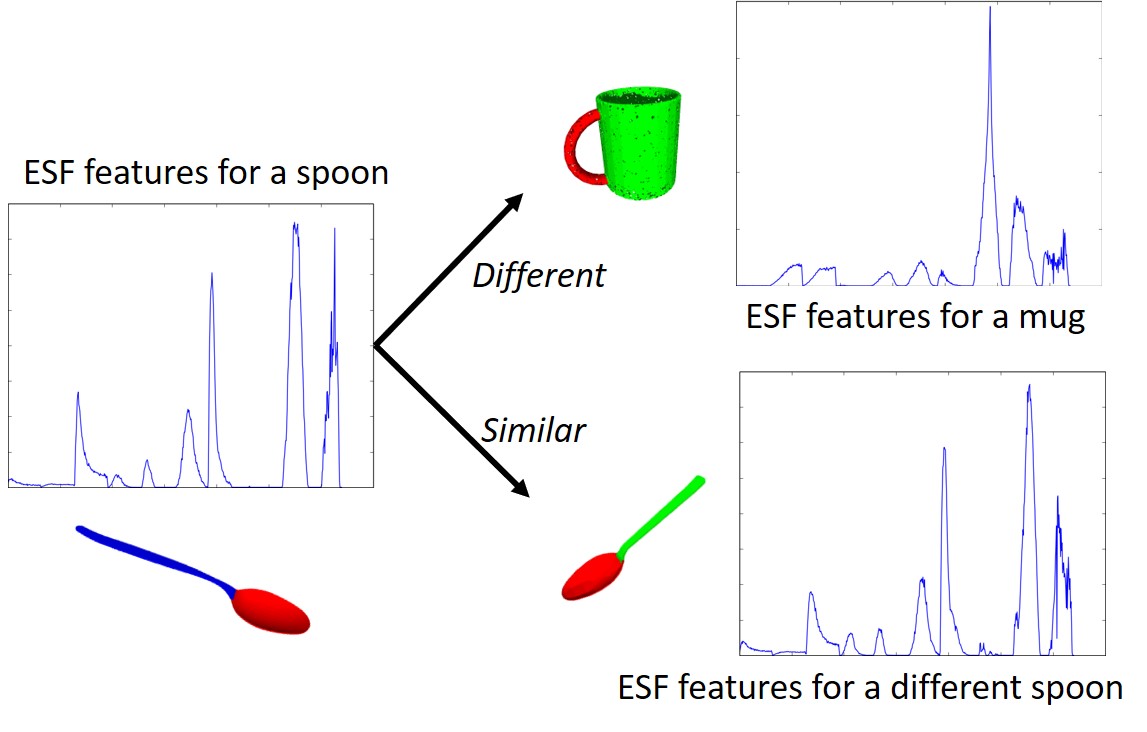}
	\captionsetup{width=\linewidth}
	\caption{The ESF features extracted for two spoons and a cup, showing the similarities of corresponding ESF features.}
	\label{fig:ESF}
\end{figure}

Prior work in tool substitution has used visual similarities between tools to identify good substitutes. Abelha et al. \cite{abelha2016model} use Superquadrics to perform tool substitution. Superquadrics (SQs) are geometric shapes that includes quadrics, but allows for arbitrary powers instead of just power of two. In their approach, the candidate tool parts are modeled using SQs and the parameters of the SQs are compared to the desired parameters of the tool for which a replacement is sought. Hence, their approach takes a reference tool as input and identifies substitutes that closely match. Schoeler et al. \cite{schoeler2016bootstrapping} learn function-shape correspondence using supervised learning and identify substitutes for a given tool using a per-part shape similarity matching. To model the tools, they use existing point cloud shape representations such as Ensemble of Shape Functions (ESF) \cite{wohlkinger2011ensemble}. ESF is a descriptor consisting of 10, 64-bin sized histograms (640-D vector), describing the shape of a point cloud (shown in Figure \ref{fig:ESF}), with demonstrated success in representing partial point clouds \cite{wohlkinger2011ensemble, Chernova-RSS-19}. In contrast to prior work, we use ESF to represent the \textit{full} tool, as opposed to individual tool parts, and use dual neural networks to match shape similarities.


\subsection{Material Classification and Matching}
Material classification involves identifying the material class of a query object. While most approaches seek to identify material classes from images \cite{schwartz2018recognizing, hu2011toward, bell2015material}, in more recent work, Erickson et al. \cite{erickson2019classification} use spectral data obtained from a hand-held spectrometer for material classification, with a validation accuracy of 94.6\%. They noted that generalizing posed a greater challenge, achieving an accuracy of 79\% on previously unseen objects. Nevertheless, spectral data helps offset some critical deficiencies of vision-based approaches, such as sensitivity to light and viewing angle. In this work, we use spectral data from a hand-held spectrometer to perform material matching. 

In contrast to material classification, material matching involves matching the similarity between two materials. Some prior work has looked at matching similarity between material compositions, predicting crystal structures \cite{yang2013data}. Other approaches seek to match industrial materials, based on various industrial parameters for describing material properties \cite{manohar2011study}. Their approach uses an existing database with extensive descriptions of industrial materials, which can be challenging to procure in a household setting. More recent work has looked at matching visual similarity between materials from a series of rendered images \cite{lagunas2019similarity}. In contrast to using vision, our approach uses spectral data to match material similarity, enabling us to capture finer nuances of material properties that often escape vision.

\subsection{Dual Neural Networks}
Dual neural networks consist of two identical networks, each accepting a different input, combined at the end with a distance metric. The parameters of the twin networks are tied and the distance metric computes difference between the final layers of the twin networks. Prior work has used dual networks extensively for matching images, with much success \cite{koch2015siamese, bromley1994signature, schroff2015facenet}. We posit that we can extend dual networks to match shapes (using ESF), and match materials (using spectral data). For both shape reasoning and material reasoning, our approach is similar to FaceNet \cite{schroff2015facenet}, in that we learn an embedding from the training data, which is then used to match a query input, for computing a similarity score. 

\section{OVERVIEW OF TOOL SUBSTITUTION FRAMEWORK}

\begin{figure}[t]
	\centering
	\includegraphics[width=0.48\textwidth]{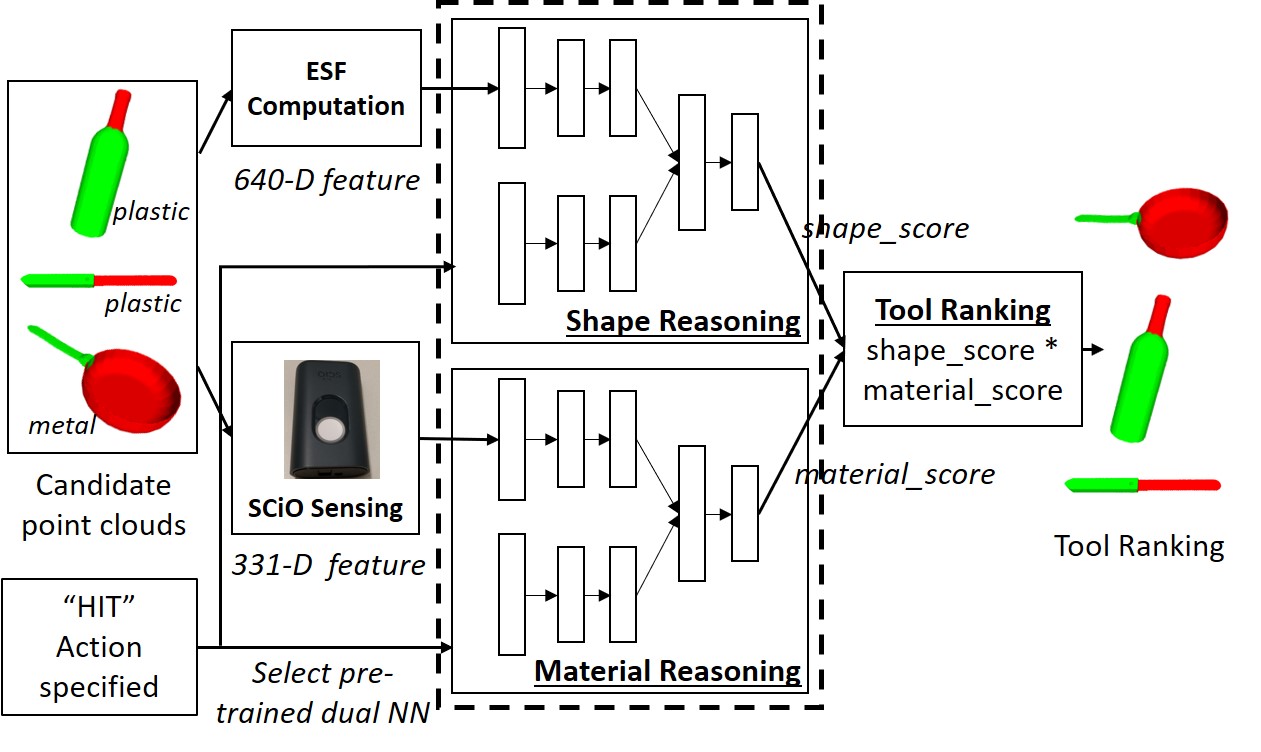}
	\captionsetup{width=\linewidth}
	\caption{The desired action and the ESF and spectral readings for each candidate, are passed through dual networks for shape and material reasoning. The tools are ranked using the combined score (product of shape and material scores).}
	\label{fig:pipeline}
\end{figure}

In this section, we present our tool substitution approach (Figure \ref{fig:pipeline} shows an overview). The research problem explored in this paper is as follows:

\textit{Given an action and a set of candidate tools, can the robot reason about shape and material of the candidate tools, to identify the best tool for performing the specified action?}

Each candidate tool in the given set, has an associated ESF feature that is computed, and a spectral scan obtained from the spectrometer. For shape reasoning, we seek to score the shapes of the candidate tools on the degree to which it matches the shapes of canonical/normative tools often used for performing the action (the canonical tool models are obtained from existing sources, such as ToolWeb \cite{abelha2016model}). Similarly, for materials, we seek to score the similarity of candidate tool materials to the desired material. For both shape and material scoring, we use supervised learning with dual neural networks. The networks are trained on pairs of inputs that are of the same/different classes, to discriminate between the class identity of the input pairs. Once the network weights are learned, we use positive examples from the training data to learn an \textit{embedding}, that acts as an \textit{anchor} for matching query inputs, similar to FaceNet \cite{schroff2015facenet}. For instance, in the case of face matching, the network is provided with an anchor image of a person's face, and is then tasked with matching the query input to the anchor image \cite{schroff2015facenet}. Here, we use the embedding as our anchor input, since it is representative of all the positive training samples (i.e., all canonical tools) in our dataset. This enables us to match the query tool to the variety of canonical shapes/materials that facilitate an action, rather than conforming to the parameters of a specific tool, as in prior tool substitution work \cite{abelha2016model, schoeler2016bootstrapping}.

\subsection{Shape Matching and Scoring}

\begin{figure}[t]
	\centering
	\includegraphics[width=0.48\textwidth]{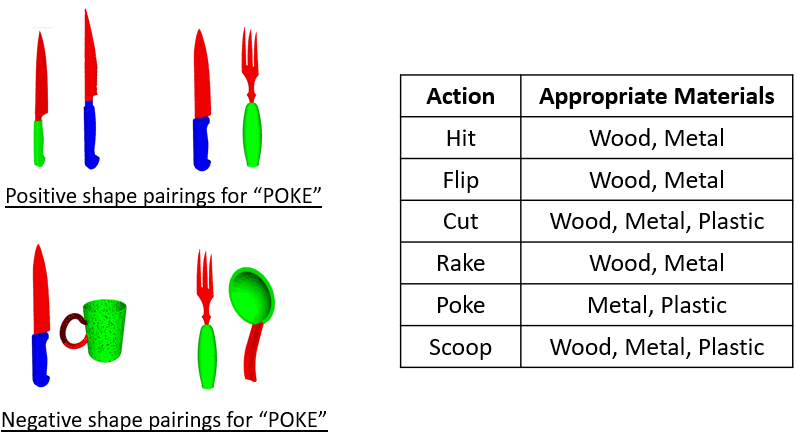}
	\captionsetup{width=\linewidth}
	\caption{Examples of positive and negative pairings for training the dual networks.}
	\label{fig:material}
\end{figure}

Shape scoring takes in a candidate tool point cloud and target action, and outputs a score indicating the degree to which the given point cloud is appropriate for the specified action. Given a set of actions $A$, a given candidate tool can be appropriate for multiple actions, e.g., a fork could be used for both poking and cutting. Hence, instead of using a single dual network trained for all the actions, we train \textit{separate} networks for each action in $A$. This also allows new actions to be trained and incorporated into the framework without affecting existing models. Given that $|A| \approx$10 for most household robots \cite{agostini2015using, tamosiunaite2019cut}, our approach can easily scale to such domains. 


\subsubsection{Feature Representation}
We represent the candidate tool point clouds using Ensemble of Shape Functions \cite{wohlkinger2011ensemble}, which captures the shape of point clouds in a 640-D vector. We use ESF as input features to our shape network.

\subsubsection{Network Architecture}
Our architecture consists of three hidden layers of 100, 100 and 25 units each. We apply tanh activation and a dropout of 0.5, after each layer. The final layer is a sigmoid computation over the element-wise $L_2$ difference between the third layer of each of the two networks, which yields the final output\footnote{Code available at: https://github.com/NithinShrivatsav/Tool-Substitution-with-Shape-and-Material-ReasoningUsing-Dual-Neural-Networks.git}. We use Adam optimizer with learning rate of 0.0001.


\subsubsection{Training}
To train\footnote{Our models are trained in Keras using Tensorflow Backend.} the dual neural network, we compiled a dataset of 3D tool models from existing online sources, namely ToolWeb \cite{abelha2016model} and 3DWarehouse. For each action, we create random pairings of tools that, based on their shape, can both be used to perform the same action and pairs that cannot (see figure \ref{fig:material}). Let $N$ be the set of training samples, then we assume that a pair $(x_i, x_j)$ is positive i.e., $y(x_i, x_j) = 1$, if both $x_i$ and $x_j$ can perform the same action and negative i.e., $y(x_i, x_j) = 0$, when either $x_i$ or $x_j$ is not suited for the action. We minimize the standard regularized binary cross-entropy loss function as:
\begin{align*}
\mathcal{L}(x_i, x_j) = y(x_i, x_j)\log(\mathbf{p}(x_i, x_j)) + \\ (1-y(x_i,x_j))\log(1-\mathbf{p}(x_i, x_j)) + \lambda |\mathbf{w}|^2
\end{align*}
The output prediction of the final layer $L$, is given as:
\begin{align*}
    \mathbf{p} = \sigma (\mathbf{w}^T (|h_{1, L-1} - h_{2, L-1}|^2) + \beta)
\end{align*}
Where $\sigma$ denotes the sigmoidal activation function, $\beta$ denotes the bias term learned during training, and $h_{1, L-1}$, $h_{2, L-1}$ denotes the final hidden layers of the twin networks respectively. The element-wise $L_2$ norm of the final hidden layers is passed to the sigmoid function. In essence, the sigmoid function computes a similarity between the output features of the final hidden layers of the two twin networks.



Once the network is trained, we learn an embedding using the positive examples (not pairings) from our training set, $x^p_i \in N$, where $x^p_i$ is a canonical tool for the action. We denote the output of the final hidden layer, for a given input $x$ as, $f(x) = h_{1, L-1}(x)$. We pass each $x^p_i$ through one of the twin networks (since both networks are identical and their weights tied), to map each input into a $d$-dimensional Euclidean space, denoted by $f(x^p_i) \in \mathbb{R}^d$. We then compute the embedding as an average over $f(x^p_i)$, for all the positive examples $x^p_i$, where $N_p$ is the number of positive examples:
\begin{align*}
    \mathcal{E}^p_{action} = \frac{1}{N_p} \sum_{i=1}^{N_p} f(x^p_i) \ \forall \ x^p_i \in N
\end{align*}
The $d$-dimensional embedding, $\mathcal{E}^p_{action}$, is computed for each action and serves as our anchor input, matched against the query input to compute a similarity score. 

\subsubsection{Prediction}
Given the ESF feature of a candidate tool, $x^c$, we first compute $f(x^c)$, using our pre-trained model as before. Then the shape score, $p_{shape}$, is computed as follows:
\begin{align*}
    p_{shape}(x^c, action) = \sigma (\mathbf{w}^T |\mathcal{E}^p_{action} - f(x^c)|^2 + \beta)
\end{align*}
This score represents the similarity between the ESF feature of the candidate tool and the embedding, $\mathcal{E}^p_{action}$, representative of all the positive examples in the training data. 

\subsection{Material Matching and Scoring}
Material scoring takes in a spectral reading and action, and outputs a score indicating the degree to which the spectral reading is suited for the specified action. As with shape scoring, we train separate models for each action. Here, we assume that the material of the \textit{acting part} of the tool is most critical to performing the action.  As a result, we simplify our model by only considering the material of the action part, e.g., we model a knife consisting of a metal blade and plastic handle, as metal. This assumption holds for the vast majority of household tools, but could be relaxed in future work.

\subsubsection{Feature Representation}
In order to extract material features for the candidate tools, we use a SCiO sensor, which is a handheld spectrometer, shown in Figure \ref{fig:pipeline}. The SCiO scans objects, returning a 331-D vector of real values. We use the SCiO readings as input to our materials network. 


\subsubsection{Network Architecture}
Our model consists of three hidden layers of 426, 284 and 128 units each. We apply tanh activation and a dropout of 0.5 after each layer. The final layer is a sigmoid computation over the element-wise $L_1$ difference between the third layer of the two networks. We use Adam optimizer with learning rate of 0.001.  
 
\subsubsection{Training}
To train the dual neural network, we use the SMM50 dataset\footnote{Dataset available at https://github.com/Healthcare-Robotics/smm50}, which contains spectrometer readings for five classes of materials: plastic, paper, wood, metal and foam. For our work, we manually identified the most appropriate material classes for each action, also shown in Figure \ref{fig:material}. We create random pairings of spectral readings, where  both materials in the pair are appropriate for the action, or either one is not. Given a set $M$ of training samples, $y(x_i, x_j) = 1$, if both materials are appropriate for a given action (as indicated by Figure \ref{fig:material}), and $y(x_i, x_j) = 0$, if either $x_i$ or $x_j$ corresponds to an inappropriate material. That is, for ``Hit'', (metal, metal) and (metal, wood) pairings are both positive examples, whereas (metal, foam) is a negative example. Note that, each pair does not necessarily consist of the same material class. The reason is that, we would like all appropriate material classes for a given action, such as metal and wood for ``Hit'', to be mapped closer in the embedding space, than metal and foam. This allows us to overcome the variance across material classes, learning an embedding space where the desired material classes are closer in distance. Our training procedure and loss function is the same as that for the shape scoring.



We compute the $d$-dimensional embedding space $\mathcal{D}^p_{action}$ as before, using the spectral readings corresponding to appropriate materials as positive examples, $x^p_i \in M$. The computed embedding represents an aggregation of the most appropriate spectral readings in the training set for an action. 

\subsubsection{Prediction}
Given the spectral reading corresponding to a candidate tool, $x^c$, we compute $f(x^c)$ using our pre-trained model. Then, similar to shape scoring, our material score, $p_{material}$, is computed as follows:
\begin{align*}
    p_{material}(x^c, action) = \sigma (\mathbf{w}^T |\mathcal{D}^p_{action} - f(x^c)| + \beta)
\end{align*}
This score represents the similarity between the material of the candidate tool and the embedding, $\mathcal{D}^p_{action}$, representative of all the positive examples in the training data. 


\smallskip Once both the shape score and material score are computed, a final score is computed as their product, and the candidate tools are ranked from highest to lowest final scores. The highest ranked tool corresponds to the best substitute. 


\section{EXPERIMENTS AND RESULTS}
In this section, we evaluate our tool substitution approach, for six actions: ``Hit'', ``Cut'', ``Scoop'', ``Flip'', ``Poke'' and ``Rake'', with five material classes: ``Metal'', ``Wood'', ``Plastic'', ``Paper'' and ``Foam''. Our experiment seeks to validate:
\begin{itemize}
    \item \textbf{Performance of shape matching}: We evaluate our approach for shape matching on a testing dataset of previously unseen object models;
    \item \textbf{Performance of material matching}: We evaluate our approach to material matching on a test set of previously unseen spectral readings;
    \item \textbf{Performance of combined shape and material reasoning for tool substitution}: We evaluate our final approach for tool substitution, comparing it with using only shape scoring, only material scoring and random ranking baselines. We test our model on a set of partial point clouds and spectral readings of real-world objects.
\end{itemize}

We evaluate the statistical significance of our results using repeated-measures ANOVA and post-hoc Tukey's test. The `**' denotes a statistically significant result with $p < 0.01$. 


\subsection{Performance of Shape Matching}
\begin{figure}[t]
	\centering
	\includegraphics[width=0.42\textwidth]{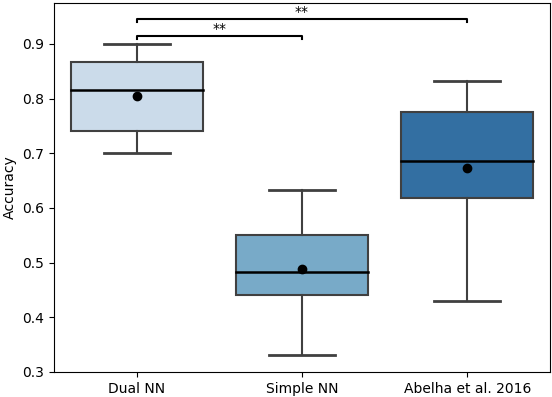}
	\captionsetup{width=\linewidth}
	\caption{Plot showing the accuracy of our shape scoring approach for each of the six actions compared to baselines.}
	\label{fig:shape_boxplot}
\end{figure}

\begin{figure}[t]
	\centering
	\includegraphics[width=0.43\textwidth]{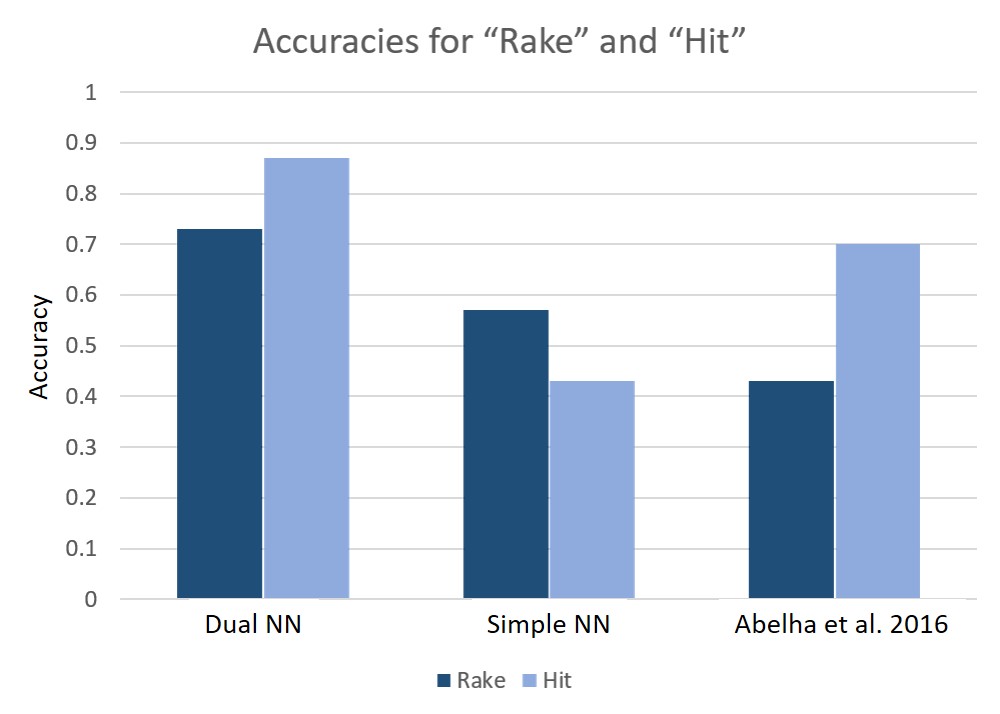}
	\captionsetup{width=\linewidth}
	\caption{Plot showing the performance of the three approaches on the hardest (Rake) and easiest (Hit) shape scoring tasks.}
	\label{fig:rake_acc}
\end{figure}

We test three different shape matching approaches on a previously unseen test set of 30 object models, collected from 3DWarehouse. The test set consists of uniquely shaped objects and allows us to measure the generalization capability of the models. We label each object in the set with the action(s) it is appropriate for (based on shape only), which acts as our ground truth label. We compare our approach (Dual NN) to a simple neural network (Simple NN) and the approach previously proposed by Abelha et al.~\cite{abelha2016model}.

For our Simple NN baseline, we evaluated multiple architectures and selected the best performing one. The best model architecture uses three hidden layers (100, 100 and 25 units each), with tanh activation and dropout of 0.5 after each layer, with sigmoid in the last layer. We use the binary cross entropy loss, with Adam optimizer. We train the model with the same dataset used for the dual networks for shape matching except, we consider individual training samples $x_i \in N$, where for any training sample $x_i$, $y(x_i) = 1$ if $x_i$ is appropriate for the action.

Our results are shown in Figure \ref{fig:shape_boxplot}. Using only shape information, we find that our approach outperforms the baselines, with an average accuracy of 81\% on the testing dataset. This shows that our network is able to generalize well to previously unseen and uniquely shaped objects, by identifying salient features that make them appropriate for a given action. Abelha et al. \cite{abelha2016model} performs reasonably well using SQs (accuracy of 67\%) but Simple NN performs poorly (49\%).  Shown in Figure \ref{fig:rake_acc}, we note that it is difficult to model objects such as rakes accurately using SQs, owing to the toothed structure of these objects. But SQs perform well on more regularly shaped objects such as hammers. We also note that the overall computation time for Dual NN and Simple NN is on average, 1.967 s and 1.911 s respectively, whereas SQ fitting with Abelha et al. \cite{abelha2016model} takes on average 342.27 s. Thus, our approach also scales better computationally.


\textbf{Key findings:} Shape matching using dual neural networks outperforms both baselines, and is computationally faster than using SQ modeling.

\subsection{Performance of Material Matching}
\begin{figure}[t]
	\centering
	\includegraphics[width=0.42\textwidth]{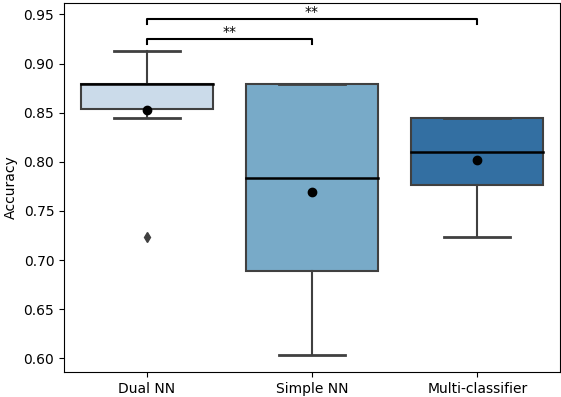}
	\captionsetup{width=\linewidth}
	\caption{Performance accuracy of our material scoring approach for each of the six actions, compared to baselines.}
	\label{fig:mat_boxplot}
\end{figure}

\begin{figure}[t]
	\centering
	\includegraphics[width=0.4\textwidth]{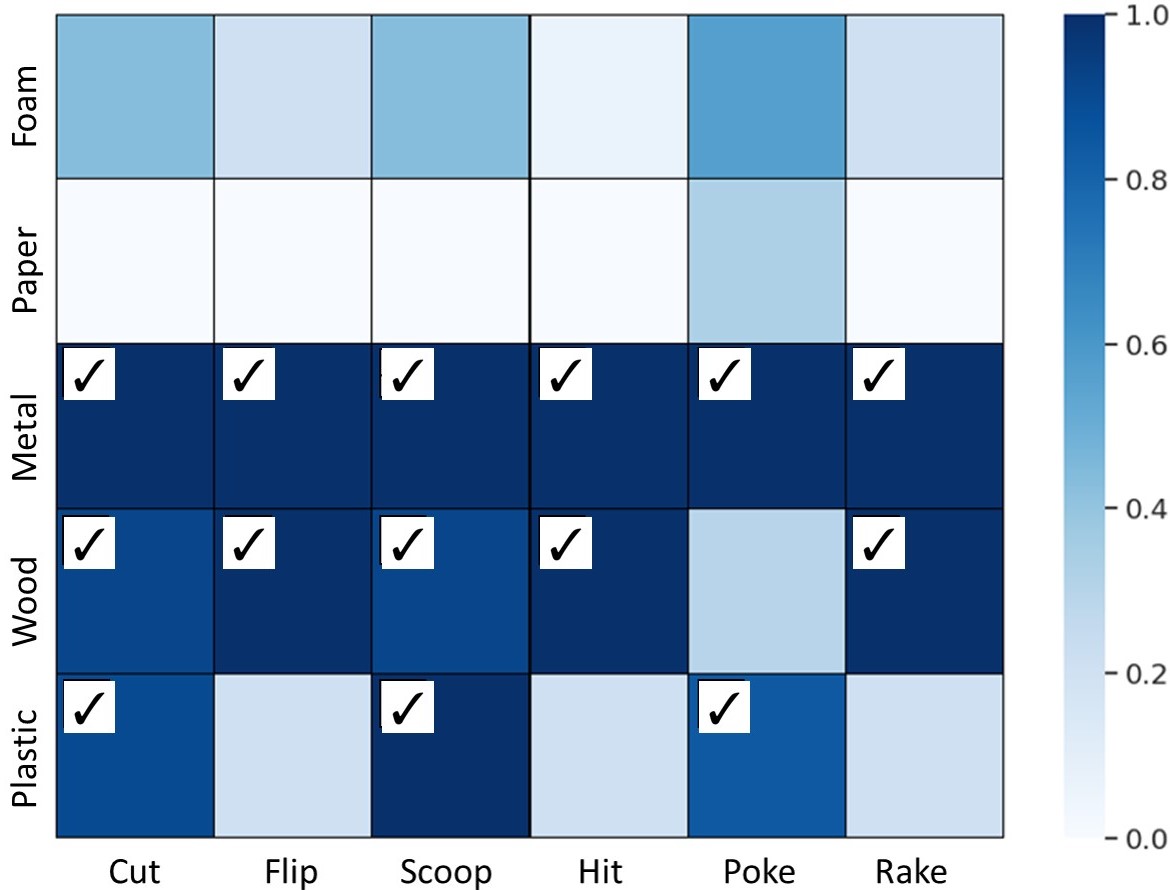}
	\captionsetup{width=\linewidth}
	\caption{Plot showing the proportion of materials predicted by the dual network for each action. Checkmarks indicate the materials appropriate for each action (best viewed in color).}
	\label{fig:conf_dNN}
\end{figure}

We test three different material scoring approaches on a test set of 58 spectral readings (14 foam, 10 metal, 3 paper, 12 wood and 19 plastic) scanned from previously unseen objects belonging to the five different material classes. Each spectral reading is labeled with the action(s) they are appropriate for, corresponding to their material class. We compare our approach (Dual NN) to a simple neural network (Simple NN). We also compare our approach to material classification previously proposed by Erickson et al. \cite{erickson2019classification} (Multi-classifier), which is used to predict the class corresponding to a reading, and then matched with our ground truth in figure \ref{fig:material}. 

Our simple neural network baseline for material matching uses four hidden layers (64, 64, 32, 32 units), with Leaky ReLU activation and dropout of 0.5. We train the model with SMM50 with training samples $x_i \in \mathcal{M}$, where $y(x_i) = 1$ if $x_i$ is an appropriate material for the action. 

Our results, Figure \ref{fig:mat_boxplot}, show that our approach outperforms the baselines, with an average accuracy of 85\%. While material classification performs almost as well (79.8\% accuracy), dual NN is able to capture the \textit{degree} of similarity between candidate and canonical materials, in contrast to only material classification. This allows the dual NN to compute a similarity score, which is beneficial when ranking tool substitutes. Shown in Figure \ref{fig:conf_dNN} is a detailed breakdown of our approach on the different classes. The darker shading denotes the proportion of materials correctly identified for that action by the model. We see that the network largely predicts suitable materials for each action, with some exceptions, such as some foam and paper objects predicted for poke. 






\textbf{Key findings:} Material matching using dual networks outperforms our baselines, and it outputs the degree of material suitability for an action, unlike material classification. 


\subsection{Performance of Combined Shape and Material Reasoning}
\label{sec:expt3}

\begin{figure}[t]
	\centering
	\includegraphics[width=0.44\textwidth]{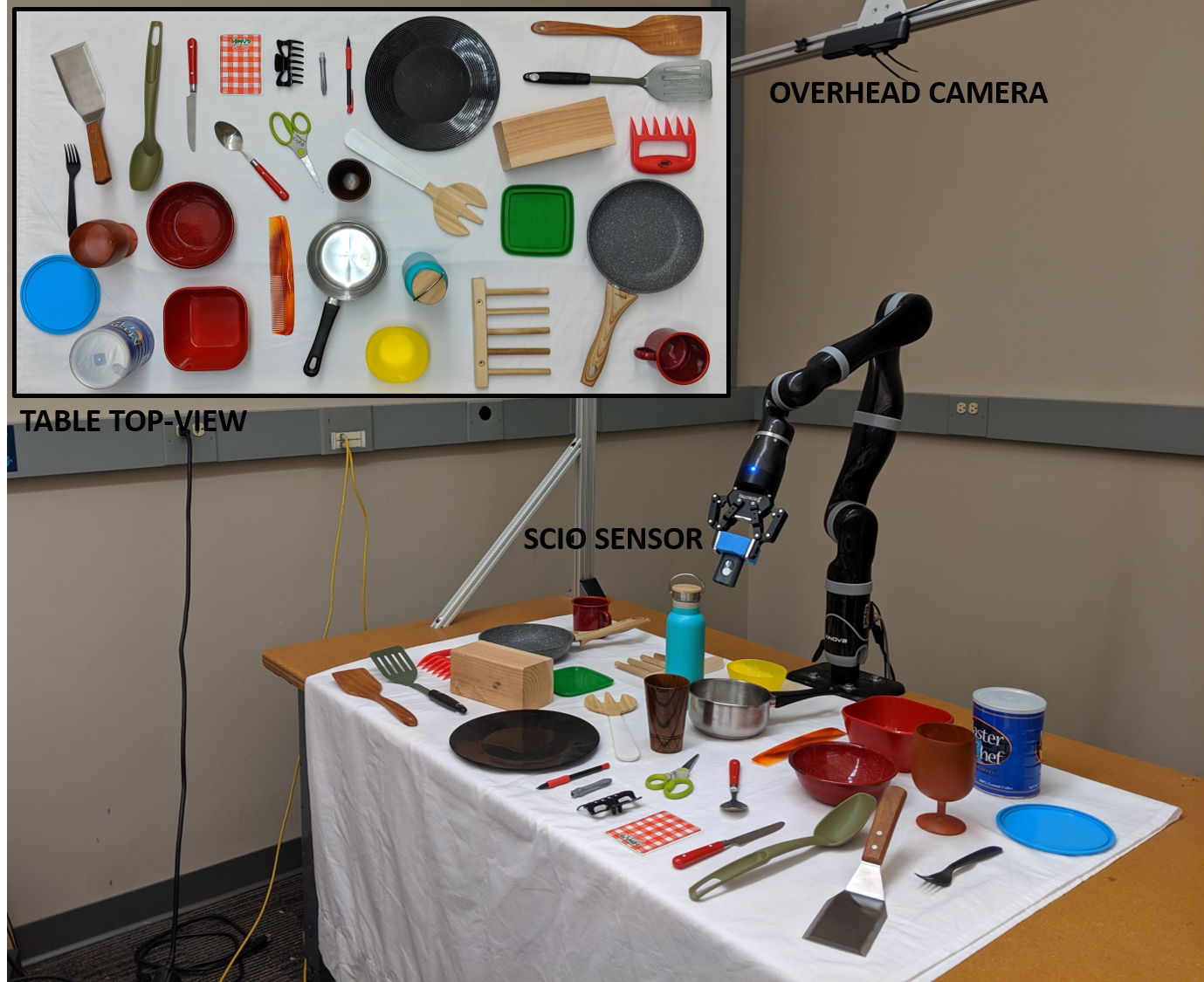}
	\captionsetup{width=\linewidth}
	\caption{Experimental setup: Shows the set of 30 objects used in experiment 3, Sec \ref{sec:expt3}, along with a sample setup of the workspace with the robot shown holding the SCiO sensor.}
	\label{fig:robot_setup}
\end{figure}

\begin{figure}[t]
	\centering
	\includegraphics[width=0.49\textwidth]{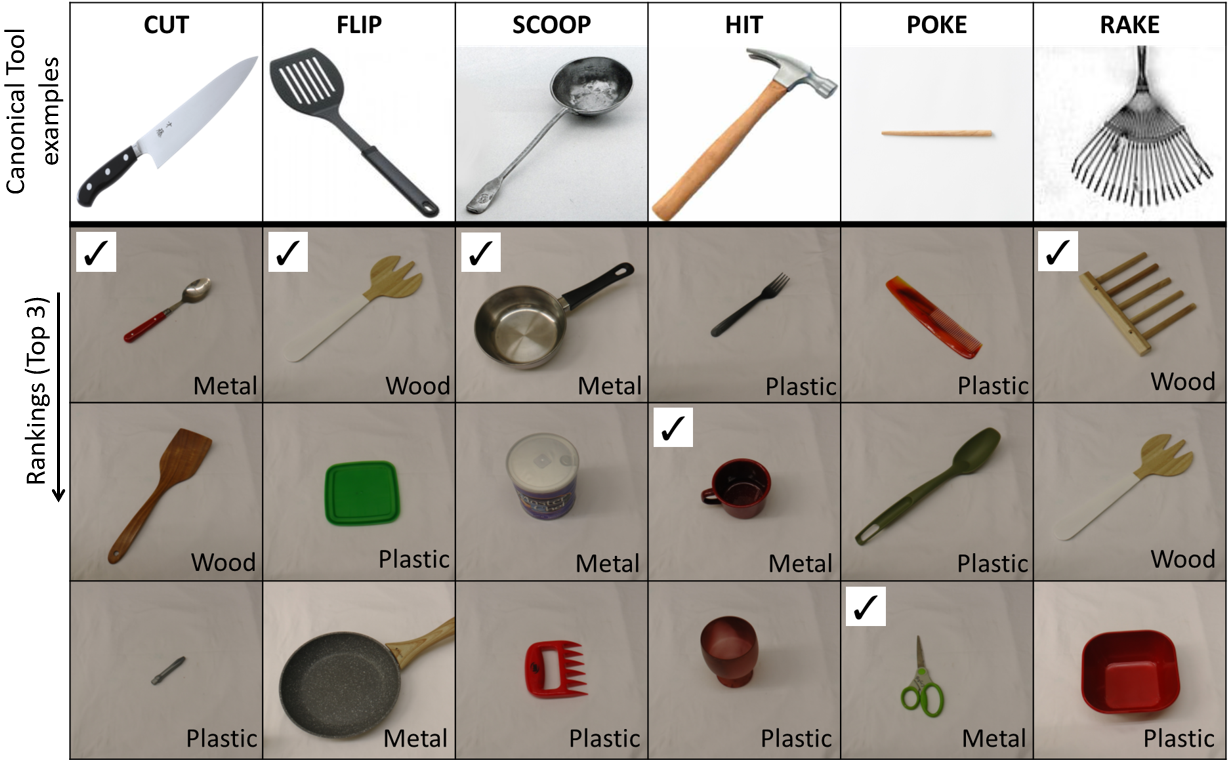}
	\captionsetup{width=\linewidth}
	\caption{First row shows examples of some canonical tools for each action. Following rows show the ranking of objects (top 3) for some of the sets. Check marks indicate the correct outputs. The actual materials of the objects are also noted.}
	\label{fig:collage}
\end{figure}

In this section, we compare the performance of dual networks for shape scoring only, material scoring only, combined shape and material scoring, and random ranking.  Unlike previous tasks, which used 3D object models from 3DWarehouse, in this task we utilize real robot data.  Our experimental setup is shown in Figure \ref{fig:robot_setup}. 3D object scans are collected using an overhead RGBD camera, and material readings are collected by the robot using the hand-held SCiO sensor. Note that the substitution task is significantly more challenging in this real-world setting because only \textit{partial} point clouds can be obtained from the overhead camera and we use single spectral scans for each object, collected by a 7-DOF robot arm.  Additionally, some object materials, such as stainless steel, were not included in SMM50 training data.

\begin{table}[t]
	\centering
	\includegraphics[width=0.48\textwidth]{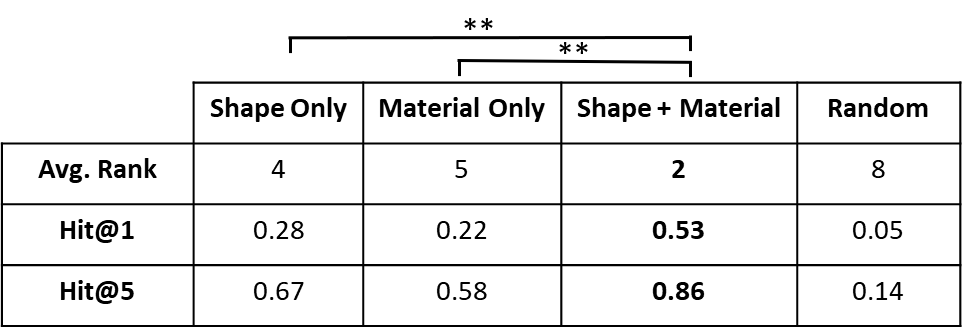}
	\captionsetup{width=\linewidth}
	\caption{Combined shape and material scoring performs better overall. Note that lower rank (min 1) and higher hit@5, hit@1 (max 1) are preferred.}
	\label{fig:results_final}
\end{table}

The 30 objects used are also shown in Figure \ref{fig:robot_setup}. For validation, we created six sets of 10 objects per action (total 36 sets).  Each set consisted of one ``correct'' substitute for the given action, and nine incorrect, which acts as our ground truth\footnote{The correct substitute was determined by three independent evaluators (with a Cronbach's alpha of 0.93).}. Given that our tool substitution approach output a ranking of candidate tools, our metrics included ``Hit@1'', indicating the proportion of sets for which the correct tool was ranked at 1; ``Average Rank'', which is the average rank of the correct tool across the test sets; and ``Hit@5'', indicating the number of times the correct tool was ranked within the top 5 ranks of our output. 


Our results are shown in Table \ref{fig:results_final}. We found that overall, our approach combining shape and material outperformed the other conditions, with an average ranking of 3 across all the sets. In particular, we note that combining shape and material significantly improved hit@5 (86\% vs 67\% for shape and 58\% material only) and hit@1 (53\% vs 28\% for shape and 22\% material only). All three approaches performed significantly better than random ranking of the objects. While our results indicated that combining shape and material reasoning improved the performance of the tool substitution pipeline, its practical application remains a significant challenge, as indicated by the low hit@1. We note that using only shape information performed better than using only material information. Our results using shape scoring only is interesting, since our original network was only trained on 3d models, yet it was able to generalize/transfer fairly well to real-world partial point clouds with 67\% hit@5. In contrast, we found that the spectral scans extracted by the robot posed a bigger challenge to the generalization of our material scoring system. This reflects the findings previously reported by Erickson et al. \cite{erickson2019classification}, and indicate that incorporating visual and haptic modalities may help improve performance. 



Figure \ref{fig:collage} shows some of the ranked substitutes returned by combined shape and material reasoning, for some of the test sets. The results highlight the challenges of working with partial RGBD data and previously unseen material scans. For example, the (closed) metal can ranked as the \#2 substitute tool for scooping is ranked highly, because its reflective surface resulted in a point cloud that resembled a concave bowl. Further, an incorrect material prediction for the metal mug, resulted in it being ranked as \#2 substitute for hitting.

\textbf{Key findings:} Combined shape and material reasoning leads to significantly improved performance for tool substitution, when compared to material or shape only.

\section{CONCLUSIONS AND FUTURE WORK}
In this work, we have contributed a novel approach to tool substitution, combining shape and material reasoning. We also presented a novel approach to using dual neural networks for performing shape matching and material matching. We evaluated our approach on six actions and our results demonstrated that our approach is able to effectively match shape and materials, with improved performance  on real-world objects, by combining shape and material reasoning. 

In our future work, we would like to address existing limitations of our approach, particularly on real world data, for partial point clouds and single spectral scans. We aim to incorporate other forms of reasoning such as haptic or visual feedback and using point cloud completion \cite{figueiredo2017automatic}, to improve performance. We also seek to reason about objects made of non-homogeneous materials, focusing on the materials of different parts of the tool as opposed to only the acting part. 

\section{ACKNOWLEDGMENTS}
This work is supported in part by NSF IIS 1564080 and
ONR N000141612835.

\addtolength{\textheight}{-12cm}   





\bibliographystyle{./IEEEtran}
\bibliography{references}

\end{document}